\documentclass[10pt,twocolumn,letterpaper]{article}

\usepackage[pagenumbers]{cvpr} 

\usepackage{graphicx}
\usepackage{amsmath}
\usepackage{amssymb}
\usepackage{booktabs}
\usepackage{url}

\usepackage[pagebackref,breaklinks,colorlinks]{hyperref}

\usepackage[capitalize]{cleveref}
\crefname{section}{Sec.}{Secs.}
\Crefname{section}{Section}{Sections}
\Crefname{table}{Table}{Tables}
\crefname{table}{Tab.}{Tabs.}

\def\cvprPaperID{*****} 
\def\confName{CVPR}
\def\confYear{2023}

\begin{document}

\title{Restormer-Plus for Real World Image Deraining: One State-of-the-Art Solution to the GT-RAIN challenge (CVPR 2023 UG2+ Track 3)}

\author{Chaochao Zheng \qquad Luping Wang$^{\ast}$ \qquad Bin Liu\thanks{Corresponding authors}\\
    Research Center for Applied Mathematics and Machine Intelligence\\
    Zhejiang Lab, Hangzhou 311121, China\\
    {\tt\small \{zhengcc, wangluping, liubin\}@zhejianglab.com}
}

\maketitle

\begin{abstract}
    This technical report presents our Restormer-Plus approach, which was submitted to the GT-RAIN Challenge (CVPR 2023 UG$^2$+ Track 3). Details regarding the challenge are available at \url{http://cvpr2023.ug2challenge.org/track3.html}. Restormer-Plus outperformed all other submitted solutions in terms of peak signal-to-noise ratio (PSNR), and ranked 4th in terms of structural similarity (SSIM). It was officially evaluated by the competition organizers as a runner-up solution. It consists of four main modules: the single-image de-raining module (Restormer-X), the median filtering module, the weighted averaging module, and the post-processing module. Restormer-X is applied to each rainy image and built on top of Restormer. The median filtering module is used as a median operator for rainy images associated with each scene. The weighted averaging module combines the median filtering results with those of Restormer-X to alleviate overfitting caused by using only Restormer-X. Finally, the post-processing module is utilized to improve the brightness restoration. These modules make Restormer-Plus one of the state-of-the-art solutions for the GT-RAIN Challenge. Our code can be found at \url{https://github.com/ZJLAB-AMMI/Restormer-Plus}.
\end{abstract}

\section{Introduction}
\label{sec:intro}

In this technical report, we provide a brief introduction to our Restormer-Plus approach submitted to the GT-RAIN Challenge (CVPR 2023 UG$^2$+ Track 3). A detailed technical report of this challenge can be found at \cite{zhang2024gt}. Our approach ranked first in terms of peak signal-to-noise ratio (PSNR) and fourth in terms of structural similarity (SSIM), see \cite{gtrain_challenge}. It was officially evaluated by the competition organizers as a runner-up solution.

Restormer-Plus is built on Restormer~\cite{restormer}. Based on our experience with the challenge, we identified 3 potential shortcomings of Restormer:
\begin{itemize}
\item poor ability to restore details and texture in rainy images,
\item vulnerability to overfitting during training,
\item suboptimal restoration of brightness in rainy images.
\end{itemize}

To address these limitations, we developed Restormer-Plus, which is presented in Section~\ref{sec:methods}. We provide a brief description of our experimental setting in Section~\ref{sec:ex_set}, and show key experimental results in Section~\ref{sec:results}.
\section{Our Method: Restormer-Plus}
\label{sec:methods}
\begin{figure}[htbp]
  \centering
   \includegraphics[width=\linewidth]{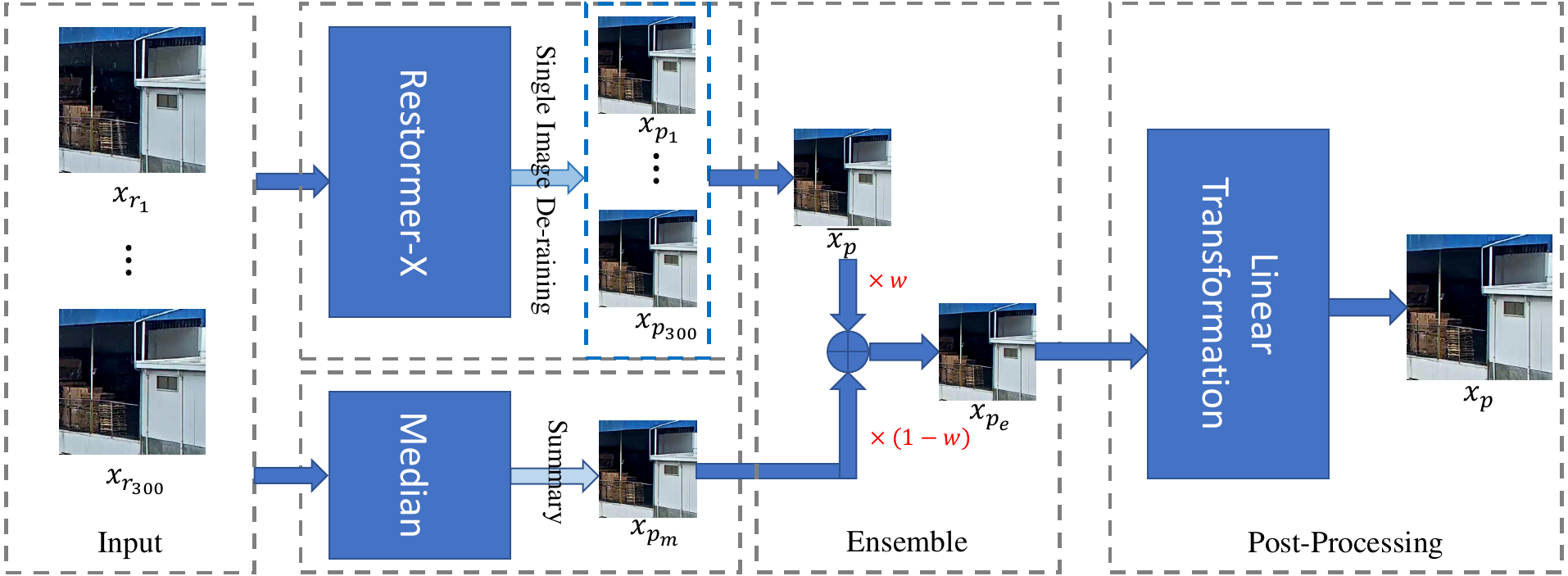}
   \caption{Overview of our de-raining solution.}
   \label{fig:overview}
\end{figure}

Restormer-Plus consists of four core modules: the single image de-raining module, the median filtering module, the weighted averaging module, and the post-processing module. An overview of Restormer-Plus is shown in Figure~\ref{fig:overview}. In the following sections, we provide a detailed introduction to each module.

\subsection{The single image de-raining module}
\begin{figure}[htbp]
  \centering
   \includegraphics[width=\linewidth]{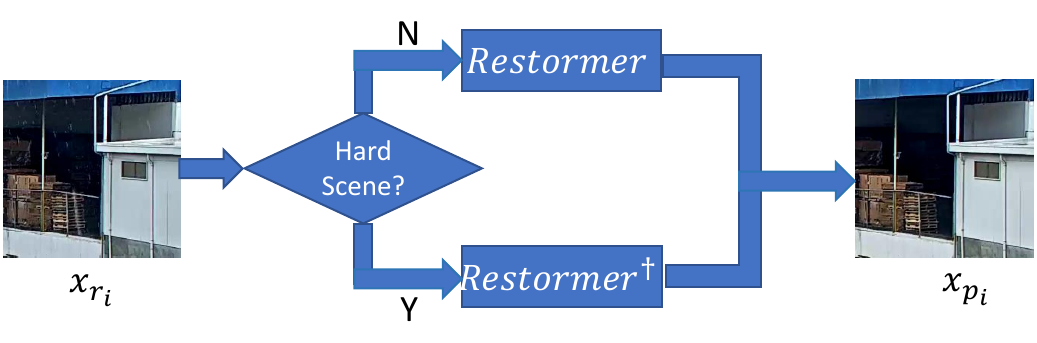}
   \caption{Illustration of Restormer-X.}
   \label{fig:restormer_x}
\end{figure}

The single image de-raining module, called Restormer-X, includes two versions of Restormer~\cite{restormer}: the basic version, referred to as $Restormer$, and the modified version, referred to as $Restormer^{\dagger}$. Figure~\ref{fig:restormer_x} illustrates Restormer-X. It is worth noting that the two versions of Restormer are trained separately and independently.

$Restormer^{\dagger}$ is built to enhance the expressive power of $Restormer$ by modifying its output layer using Equation~(\ref{eq:restormer_p}).
\begin{equation}
\label{eq:restormer_p}
    x_{p_{i}} = f_w(v_{x_{r_{i}}}) x_{r_{i}} + f_b(v_{x_{r_{i}}})
\end{equation}
where $x_{r_{i}}$ denotes the $i^{th}$ noisy image and $v_{x_{r_{i}}}$ is its feature vector, i.e., the output of the \textbf{Refinement} block of $Restormer$. The functions $f_w$ and $f_b$, which represent the weight and bias of the output layer, respectively, have the same structure as \textbf{Conv2d} but with separate trainable variables. Clearly, $Restormer^{\dagger}$ has a stronger expressive power than $Restormer$, making it better suited for challenging hard scenes.

\begin{figure}[htbp]
  \centering
   \includegraphics[width=\linewidth]{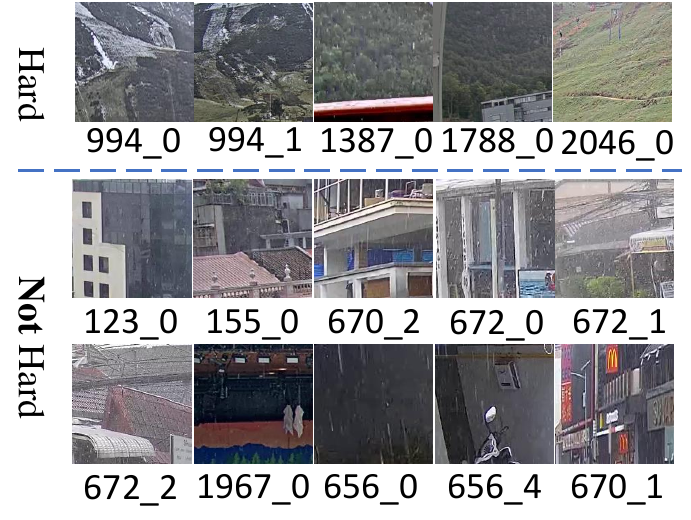}
   \caption{Scene split of GT-Rain test dataset.}
   \label{fig:scene_split}
\end{figure}

Before using Restormer-X to perform inference on each scene, we need to determine whether the scene is hard to process. Figure~\ref{fig:scene_split} shows the results of scene splitting for the GT-Rain test dataset. In the figure, scenes that are considered ``hard" contain drizzling rain and/or complex textures, which intuitively require a stronger expressive power output layer rather than a basic residual output layer.

\subsection{The median filtering module}
This module utilizes a fundamental prior knowledge that raindrops will appear at different spatial positions at different times, while the scene background remains unchanged over time. Based on this prior knowledge, we only need to perform median filtering on each pixel of the 300 noisy images in a scene along the temporal dimension, and the resulting output is used as the restored image.
Mathematically, this module can be formulated as follows:
\begin{equation}
\label{eq:rule}
    x_{pm} = \mbox{Median}(x_{r_{1}},x_{r_{2}},\ldots,x_{r_{T}}),
\end{equation}
where $T=300$.
\subsection{The weighted averaging module}
As mentioned earlier, the mechanism for restoring rainy images in Restormer-X differs significantly from that of the median filtering module. In principle, Restormer-X performs fine-grained restoration on the noisy image in a supervised manner, which is prone to overfitting. To address this issue, we take a weighted average of the output results from Restormer-X and the median filtering module outputs, thus mitigating or avoiding overfitting.

Specifically, firstly, we average the results of each scene finely restored by Restormer-X as follows
\begin{equation}
   \overline{x_{p}}=\mbox{Mean}(x_{p_{1}},x_{p_{2}},\ldots,x_{p_{T}}),
\end{equation}
where $T=300$.
Then we compute the weighted average of $\overline{x_{p}}$ and $x_{pm}$, which is given by Equation (\ref{eq:ensemble}), as follows,
\begin{equation}
\label{eq:ensemble}
    x_{pe} = w \overline{x_{p}} + (1 - w) x_{pm}
\end{equation}
where $w$ represents the weight, which is determined using a backtracking method on the GT-Rain-val dataset. Ultimately, $w$ is set to 0.9.
\subsection{The post-processing module}
To adjusts the brightness of the recovered images, we perform post-processing on results yielded from the above weighted averaging module. Specifically, we perform a linear transformation on each scene independently using Equation~(\ref{eq:post_process}) as below
\begin{equation}
\label{eq:post_process}
    x_{p, i} = \hat{a} \odot x_{pe, i} + \hat{b}
\end{equation}
where $i$ denotes the pixel index, $\odot$ denotes element-wise multiplication, $\hat{a}, \hat{b} \in R^{3}$ are parameters of the linear transformation, whose values are determined by Equation (\ref{eq:ab_val}) based on a set of pixel-samples $\{(x_{pe, j}, x_{est, j})\}, j \in I$, where $I=\{i_{1}, i_{2}, \cdots, i_{K}\}$, $x_{est, j}$ is the estimated value of the $j^{th}$ pixel.
\begin{equation}\label{eq:ab_val}
    \begin{array}{ccl}
         \hat{a}_{k} &=& \frac{K\sum_{j \in I}(x_{pe, j, k}\times x_{est, j, k}) -\sum_{j \in I}x_{pe, j, k}\sum_{j \in I}x_{est, j, k}}{K\sum_{j \in I}x_{pe, j, k}^{2} - (\sum_{j \in I}x_{pe, j, k})^{2}}\\
         \hat{b}_{k} &=& \frac{\sum_{j \in I}x_{est, j, k}}{K} - \hat{a}_{k} \frac{\sum_{j \in I}x_{pe, j, k}}{K}
    \end{array}
\end{equation}
In the above equation, $k\in\{1, 2, 3\}$ represents the index of channels. The value of $K$ was set to a very small value of 10 in comparison to the total number of pixels in an image.

\begin{figure}[htbp]
  \centering
   \includegraphics[width=\linewidth]{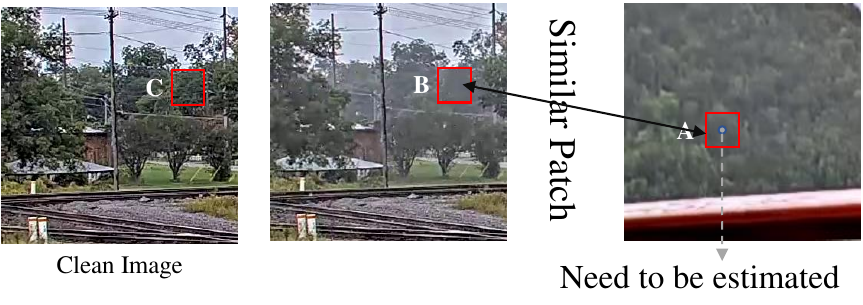}
   \caption{Illustration of how to compute $x_{est, j}$.}
   \label{fig:est_pixel}
\end{figure}

\paragraph{Computation of $x_{est, j}$} Here we describe how to compute $x_{est, j}$ as shown in Equation~(\ref{eq:ab_val}). Suppose that the third image in Figure~\ref{fig:est_pixel} is the median result of a scene in the test dataset, and that we need to estimate the value of a pixel in this image. Firstly, we select a small patch $A$ around the pixel to be estimated. We then search for a set of patches that are most similar to patch $A$ in the train and/or valid datasets. If the patch $B$ in the second image (the median result of a scene in the train dataset) is one of the most similar patches to $A$, then we take the mean value of the corresponding patch $C$ in the clean image as the estimated pixel value.

\paragraph{Post-Processing+} To improve the stability of the post-processing method introduced above, we sample several sets of pixel indices denoted as $I_{1}, I_{2}, \cdots, I_{N}$. For each set $I_{i}$, we can compute the corresponding $\hat{a}_{i, k}$ and $\hat{b}_{i, k}$ based on Equation~(\ref{eq:ab_val}) for the $k^{th}$ channel. We then take the mean value of $\hat{a}_{i, k}, \hat{b}_{i, k}, i=1,2,\cdots,N$ as the coefficients of Equation~(\ref{eq:post_process}) for the $k^{th}$ channel.
\section{Experimental setting}
\label{sec:ex_set}
\paragraph{Datasets}
We used only the GT-Rain~\cite{gtrain} dataset provided by the challenge organizer in our experiments. The dataset consists of four subsets: GT-Rain-train, GT-Rain-val, GT-Rain-test, and Streaks-Garg06. We used GT-Rain-train to train the models $Restormer$ and $Restormer^{\dagger}$ in the single image de-raining module. GT-Rain-val was used to determine the hyperparameters in our solution. GT-Rain-test was used to test our solution by submitting the restored results. Finally, Streaks-Garg06 was used for rain augmentation during the training phase of $Restormer$ and $Restormer^{\dagger}$.

\paragraph{Training of $Restormer$ and $Restormer^{\dagger}$} All the hyperparameters used for training $Restormer$ and $Restormer^{\dagger}$ were kept the same. Data augmentation techniques such as rain-augmentation, random rotation, random cropping, random flipping, and MixUp~\cite{mixup} with a probability of 0.5 were employed during training. The hyperparameters in $Restormer$ and $Restormer^{\dagger}$ were set according to~\cite{restormer}. The input patch size was fixed at 256. We used $AdamW$ as the optimizer with an initial learning rate of 3e-4 and weight decay of 1e-4. The learning rate was scheduled based on $CosineAnnealingLR$ after warming up for 4 epochs. The model was trained for a total of 20 epochs on 4 V100-GPU-32GB, with a batch size of 2 for each GPU. All related source code was implemented using PyTorch 1.12.1.
\section{Results}
\label{sec:results}
\begin{figure}[htbp]
  \centering
   \includegraphics[width=\linewidth]{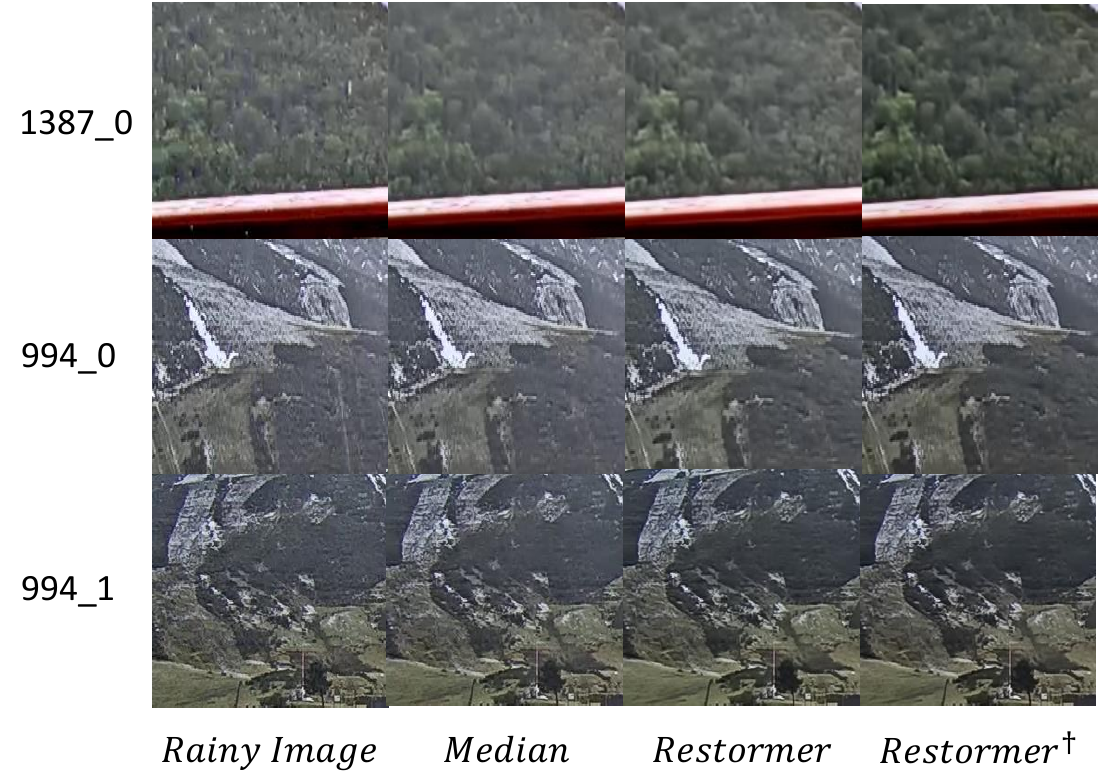}
   \caption{Comparison between $Median$, $Restormer$ and $Restormer^{\dagger}$ on some hard scenes.}
   \label{fig:restormer_comp}
\end{figure}

Figure~\ref{fig:restormer_comp} compares the restoration results of $Median$, $Restormer$ and $Restormer^{\dagger}$ on some hard scenes 1387\_0, 994\_0 and 994\_1. We can observe that $Restormer^{\dagger}$ provides restored images with more accurate color rendition, making them appear closer to the actual non-rainy scenes.

To provide a clear view of the contributions made by the ideas included in our solution, we conducted an ablation study experiment and present the results in terms of PSNR and SSIM in Table~\ref{tab:methods_res}. Our findings indicate that each designed module used in our Restormer-Plus is both effective and necessary.
\begin{table}[htbp]
  \centering
  \caption{Ablation study experiment result}
    \begin{tabular}{l|cc}
    \toprule
          & \textbf{PSNR} & \textbf{SSIM} \\
    \midrule
    \textbf{Median} & 22.502 & 0.775 \\
    \textbf{$Restormer$} & 25.218 & 0.802 \\
    \textbf{$Restormer^{\dagger}$} & 25.395 & 0.795 \\
    \textbf{Restormer-X} & 25.545 & 0.799 \\
    \textbf{+Weighted averaging} & 25.760 & 0.805 \\
    \textbf{+Post-Processing} & 26.503 & 0.822 \\
    \textbf{+Post-Processing+} & 26.633 & 0.825 \\
    \bottomrule
    \end{tabular}%
  \label{tab:methods_res}%
\end{table}%
\section{Conclusions}
We presented Restormer-Plus, our solution submitted to the GT-RAIN Challenge (CVPR 2023 UG$^2$+ Track 3). It was officially evaluated by the competition organizers as the runner-up solution. Restormer-Plus is an improved version of Restormer~\cite{restormer}, adapted to this challenge. Our work indicates that exploring and exploiting domain knowledge through operations such as prediction ensembling, data pre-processing, or post-processing is a necessary strategy for improving the performance of end-to-end deep learning algorithms in real-life scenarios, which suffer from highly scarce annotated data.
{\small
\bibliographystyle{ieee_fullname}
\bibliography{egbib}
}
\end{document}